# Solving combinational optimization problems with evolutionary single-pixel imaging

WEI HUANG,[1,2] JIAXIANG LI,[3] SHUMING JIAO[1,*] AND ZIBANG ZHANG[3]

[1]*Peng Cheng Laboratory, Shenzhen 518055, Guangdong, China*
[2]*College of Electrical and Information Engineering, Hunan University, Changsha, Hunan, China*
[3]*Department of Optoelectronic Engineering, Jinan University, Guangzhou 510632, Guangdong, China*
*\*Corresponding author: jiaoshm@pcl.ac.cn*

**Abstract: Single-pixel imaging (SPI) is a novel optical imaging technique by replacing the pixelated sensor array in a conventional camera with a single-pixel detector. In previous works, SPI is usually used for capturing object images or performing image processing tasks. In this work, we propose a SPI scheme for processing other types of data in addition to images. An Ising machine model is implemented optically with SPI for solving combinational optimization problems including number partition and graph maximum cut. Simulated and experimental results show that our proposed scheme can optimize the Hamiltonian function with evolutionary illumination patterns.**

As a novel optical imaging technique, single-pixel imaging (SPI) [1,2] has received much attention in recent years. A pixelated sensor array is usually required for a conventional camera. However, the sensor only has one single pixel in single-pixel imaging (SPI), shown in Fig. 1. A set of different illumination patterns are sequentially projected onto the object image. For each illumination, the single-pixel detector will collect the total light intensity of the object scene, which can be mathematically modelled as the inner product between the object image and one illumination pattern. Finally, a single-pixel intensity sequence is recorded after many illuminations. The object image can be computationally reconstructed when the illumination patterns and the intensity sequence are both known.

SPI is inherently designed for capturing an object image and most previous works about SPI are related to image acquisition and processing. Examples include image denoising [3,4], image classification [5,6], image recognition [7,8] and object tracking in an image [9-12]. In fact, a SPI system can also be considered as an optical machine learning system (or optical information processing system) for performing calculation tasks for other types of data, in addition to images. In this work, we employ SPI for solving combinational optimization problems based on an Ising machine model and evolutionary illumination patterns.

An Ising machine model [13-16] refers to a system consisting of discrete variables that represent magnetic dipole moments of atomic "spins" that can be in one of two states (+1 or -1). The total system energy depends on the states of all the spins. Some physical phenomena such as phase transition of two-dimensional square-lattice can be simulated by an Ising machine model. Many other combination optimization problems in biology, telecommunications, logistics, economy and social networks can also be potentially modeled by an Ising machine. Ising machine can be implemented optically in various ways such as a network of degenerate optical parametric oscillators [13], encoding of diffractive light field [14,15], and a mesh of cascaded Mach–Zehnder interferometer (MZIs) [16]. As far as the author knows, the Ising machine model has been seldom combined with SPI in previous works. Compared with the optical systems above [13-16], SPI has a simple and low cost experiment setup under incoherent lighting conditions. The objective of an Ising machine system is to achieve the ground state of an Ising Hamiltonian, which is an optimization process. In a recent work, the image reconstruction of SPI is formulated as an iterative evolutionary optimization of illumination patterns [17]. In this work, an Ising machine model is implemented by SPI in the following way to solve combinational optimization problems.

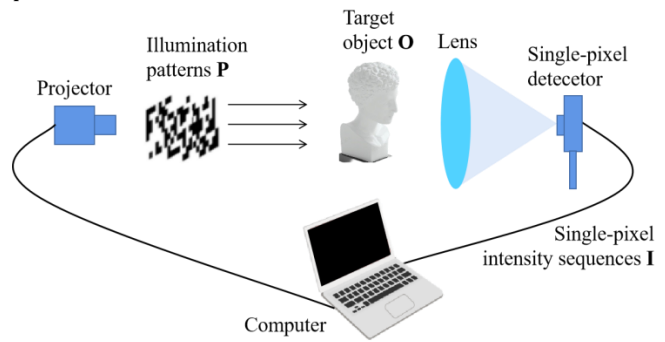

Fig. 1. Optical setup of a SPI system

It is assumed that there are totally N elements in the model and the state of each element is indicated by $\sigma_i = 1$ or $-1$ ($1 \leq i \leq$ N). There is a weighting factor $w_{ij}$ between the ith element and the jth element. The system energy can be defined by the following Hamiltonian function H, given by Equation (1). The optimal solution can be obtained by minimizing H.

$$H = -\sum_{i=1}^{N}\sum_{j=1}^{N} \sigma_i \sigma_j w_{ij} \quad (1)$$

Two typical combinational optimization problems that can be solved by an Ising machine are number partition problem and graph maximum cut problem. In the number partition problem, a set of N numbers, e.g. {1,2,5,6,7,9}, are divided into two groups and the objective is that the summations of the numbers in each group are as close as possible (e.g. {1,5,9} and {2,6,7}). It will be equivalent to an Ising model with the following settings: (1) It is indicated by $\sigma_i = 1$ or $-1$ ($1 \leq i \leq N$) for whether each number belongs to the first group or the second group; (2) For any two pairs of numbers (ith one and jth one), the weighting factor $w_{ij}$ is defined as their multiplication product.

In the graph maximum cut problem, a number of N nodes are interconnected as a network and there is a pairwise weighting $w_{ij}$ between the ith node and jth node. The objective is to divide the nodes two sub-networks and the total connection strength between two sub-networks are maximized. Fig. 2 shows an exemplar solution of graph maximum cut problem. There are 6 nodes, and the connection strength between two sub-networks is higher than that within each sub-network. Similarly, it is indicated by $\sigma_i = 1$ or $-1$ ($1 \leq i \leq N$) for whether each node belongs to the first or second sub-network. This problem is also equivalent to an Ising model. For brevity, the mathematical proofs are skipped.

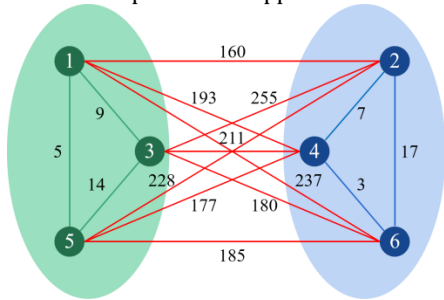

Fig. 2. Exemplar solution of graph maximum cut problem.

Our proposed optoelectronic Ising machine scheme with evolutionary SPI is described as follows, shown in Fig. 3. In SPI, the Hamiltonian function can be considered as an inner product between a map of weighting factors $w_{ij}$ and a correlation map of spinning states $\sigma_i\sigma_j$. The former one is represented by a grayscale object image with N × N pixels (pixel intensities proportional to weighting factor values) and the latter one is represented by a programmable binary illumination pattern with N × N pixels. The illumination pattern is generated from a corresponding one-dimensional vector $[\sigma_1\ \sigma_2\ \ldots\ \sigma_N]$ with length N representing the states of all the elements. The recorded single-pixel light intensity value is equivalent to the opposite of the function value since it is mathematically the inner product between the object image and one illumination pattern.

In order to finally obtain one optimal solution, a SPI scheme with evolutionary illumination patterns can be employed. It is assumed that there are totally K illumination patterns in each iteration, where K is referred to as population size. We first randomly generate K different initial state vectors $[\sigma_1\ \sigma_2\ \ldots\ \sigma_N]$ and each vector is referred to as an individual. Then corresponding K illumination patterns representing the product of each pair of $\sigma_i$ and $\sigma_j$ elements can be generated as well. After that K single-pixel values can be obtained and they are ranked from minimum to maximum. The state vectors (or illumination patterns) corresponding to first J largest single-pixel values will be preserved in the next iteration and the remaining (K-J) individuals will be replaced by crossover results of the J ones. Crossover refers to the generation of a new individual by swapping vector elements at randomly selected positions between two parent individuals. Moreover, a certain percentage of random mutation is introduced and the binary values of some vector elements are randomly modified as well. The individuals (i.e. state vectors or illumination patterns) are gradually evolved as the number of iterations increases and the maximum recorded single-pixel value will gradually increase. A converged optimal solution (or at least near-optimal solution) representing the minimum Hamiltonian function value can be obtained in most time, revealed by the optimal state vector.

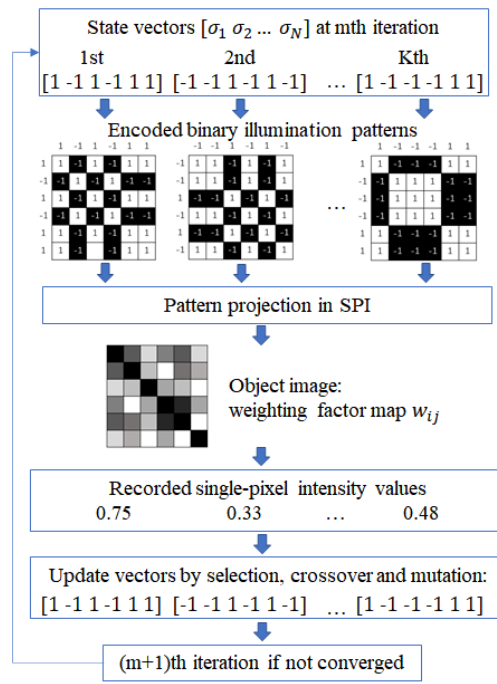

Fig. 3. Proposed Ising machine scheme with evolutionary SPI

Our proposed scheme is verified by both computer simulation and optical experiment. Three simulations are implemented for both number partition problem and graph maximum cut problem. In the number partition problem, three random number sets {2,4,5,6,9} , {1,2,3,4,5,6,7} and {1,2,5,7,9,11,15,16,18} are selected to generate the corresponding weight maps by multiplying any two elements as the object images, shown in Fig. 4(a)(c)(e) respectively. The number of individuals is K=6 and the number of iterations is 6. Using evolutionary SPI described above, the optimal solution can be obtained iteratively, and the error curves are shown in Fig. 4(b)(d)(f). Meanwhile, the partition result of every iteration is recorded as shown in Fig.5. (a)-(c), green mark is the correct classification result. Similarly, for the graph maximum cut problem, three networks with random connection strengths are selected to generate weight images, shown in Fig. 6. (a)(c)(e) respectively. The final node grouping results are {1,3,5} − {2,4,6} , {1,2,3,4} − {5,6,7,8} and {1,3,5,7} − {2,4,6,8,9}. The total cut result of every iteration is shown in Fig. 6(b)(d)(f) respectively. And the maximum cut result is shown in Fig. 7. (a)-(c). All the results above show that evolutionary SPI can obtain the optimal state vector after a few iterations.

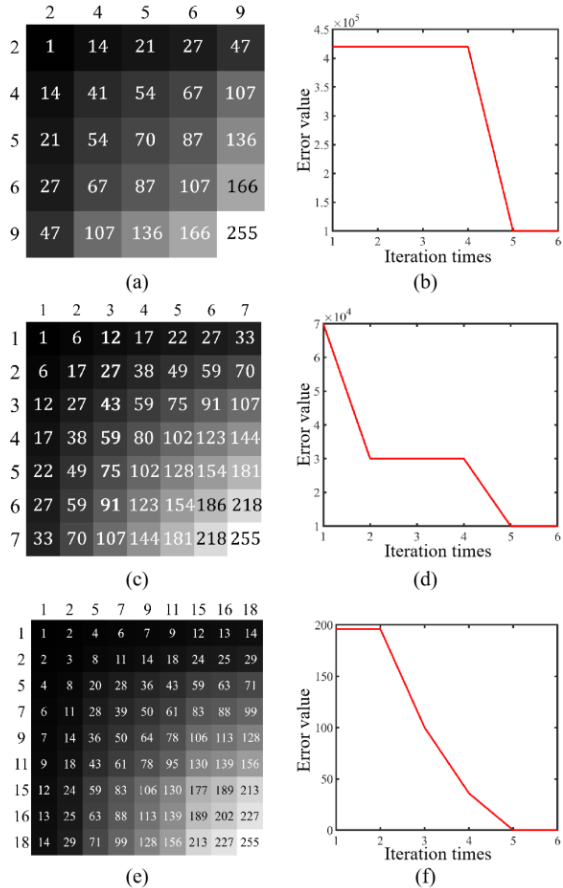

Fig. 4. Simulation results for the number partition problem: (a)(c)(e) weight images; (b)(d)(f) error curves in each iteration.

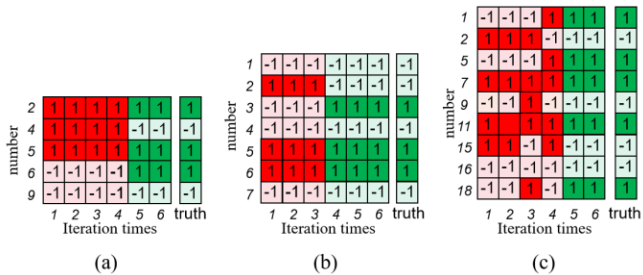

Fig. 5. Simulated optimal state vectors in each iteration for the number partition problem. (a) {2,4,5,6,9}; (b){1,2,3,4,5,6,7}; (c) {1,2,5,7,9,11,15,16,18}.

Further simulations are implemented by increasing the number of elements from N=10 to N=400 with intervals 10 for the number partition problem. The number of iterations are recorded until the correct partition results are obtained in each simulation, shown in Fig. 8. It can be seen that even if the number of elements increases to 400 and the problem is at a large scale, our proposed scheme can still perform the number partition task but require more iterations.

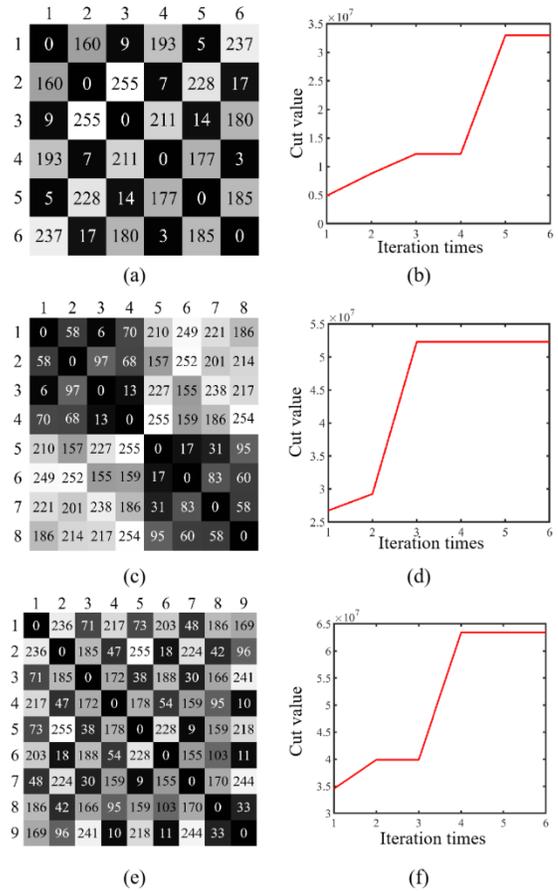

Fig. 6. Simulation results of the graph maximum cut problem: (a)(c)(e) weight images; (b)(d)(f) cut value curves in each iteration

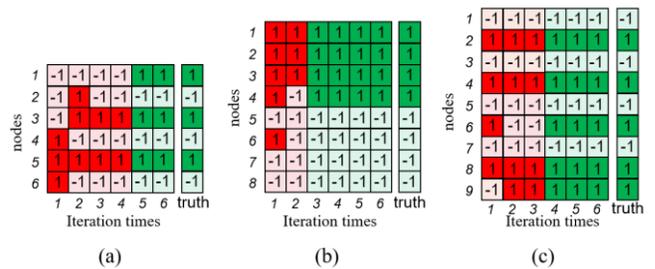

Fig. 7. Simulated optimal state vectors in each iteration for the graph maximum cut problem corresponding to (a)Fig. 6(a);(b) Fig. 6(c);(c) Fig. 6(e).

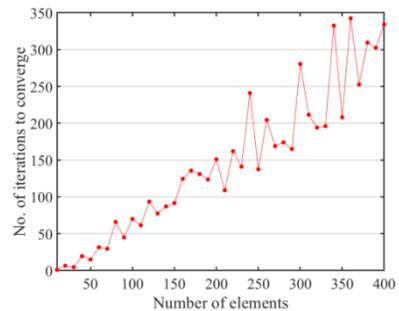

Fig. 8. Number of iterations to converge in solving number partition problem at various scales with our proposed scheme

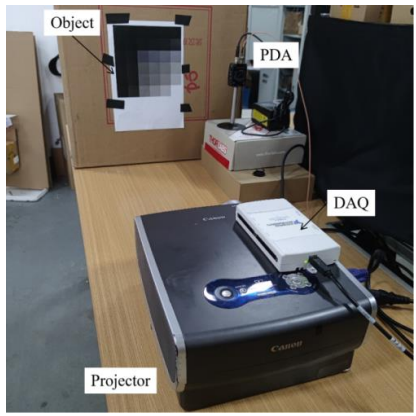

Fig. 9. Experimental SPI setup in this work

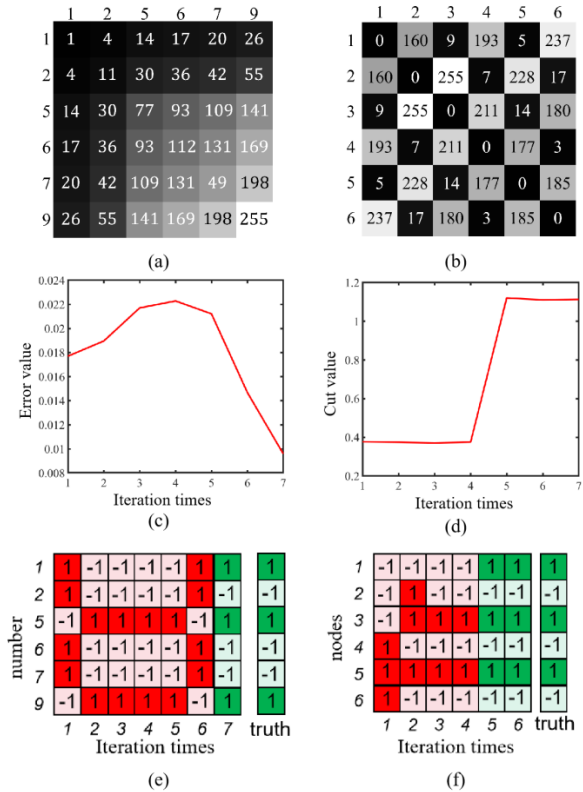

Fig. 10. Experimental results of number partition and graph maximum cut: (a) weight image for number partition; (b) weight image for graph maximum cut; (c) error value curve in each iteration for (a); (d) cut value curves in each iteration for (b); (e) optimal state vectors in each iteration for (a); (f) optimal state vectors in each iteration for (b).

The experimental SPI setup is shown in Fig. 9. The binary illumination patterns are projected onto the printed weighting factor pattern on a paper sheet by a Canon REALiS SX7 commercial projector. A Thorlabs 100A2 PDA (photo-diode array) is used as the single-pixel detector and its recorded data are collected by a NI-USB-6216 data acquisition card connected to a computer. In the experiment, one example for the number partition problem and one example for the graph maximum cut problem are tested (N=6). The total iteration number is 7 in both two experiments. The results are shown in Fig. 10. It shall be noted that the error curve of Fig. (d) has a slight rise in the middle part and it is caused by the signal acquisition error of the single-pixel detector. But the proposed scheme can still give the final correct result. After 7 iterations, a converged optimal solution emerges in all the two experiments.

In conclusion, we propose an optoelectronic Ising machine scheme with evolutionary SPI that is capable of solving combinational optimization problems such as number partition and graph maximum cut optically. The simulated and experimental results show that our proposed scheme can optimize the Hamiltonian function effectively.